\documentclass[conference]{IEEEtran}
\IEEEoverridecommandlockouts
\usepackage{cite}
\usepackage{amsmath,amssymb,amsfonts}
\usepackage{graphicx}
\usepackage{textcomp}
\usepackage{xcolor}
\usepackage{hyperref}
\usepackage{balance}

\def\BibTeX{{\rm B\kern-.05em{\sc i\kern-.025em b}\kern-.08em
    T\kern-.1667em\lower.7ex\hbox{E}\kern-.125emX}}

\begin{document}

\title{A Data-Centric Vision Transformer Baseline for SAR Sea Ice Classification}

%
%

\author{
\IEEEauthorblockN{David Mike-Ewewie}
\IEEEauthorblockA{\textit{Department of Computer Science} \\
\textit{The University of Texas Permian Basin}\\
Odessa, Texas, USA \\
mike\_d63291@utpb.edu}
\and
\IEEEauthorblockN{Panhapiseth Lim}
\IEEEauthorblockA{\textit{Department of Computer Science} \\
\textit{The University of Texas Permian Basin}\\
Odessa, Texas, USA \\
lim\_p65274@utpb.edu}
\and
\IEEEauthorblockN{Priyanka Kumar}
\IEEEauthorblockA{\textit{Department of Computer Science} \\
\textit{The University of Texas Permian Basin}\\
Odessa, Texas, USA \\
kumar\_p@utpb.edu}
}

\maketitle

\begin{abstract}
Accurate and automated sea ice classification is important for climate monitoring and maritime safety in the Arctic. While Synthetic Aperture Radar (SAR) is the operational standard because of its all-weather capability, it remains challenging to distinguish morphologically similar ice classes under severe class imbalance. Rather than claiming a fully validated multimodal system, this paper establishes a trustworthy SAR-only baseline that future fusion work can build upon. Using the AI4Arctic / ASIP Sea Ice Dataset (v2), which contains 461 Sentinel-1 scenes matched with expert ice charts, we combine full-resolution Sentinel-1 Extra Wide inputs, leakage-aware stratified patch splitting, SIGRID-3 stage-of-development labels, and training-set normalization to evaluate Vision Transformer baselines. We compare ViT-Base models trained with cross-entropy and weighted cross-entropy against a ViT-Large model trained with focal loss. Among the tested configurations, ViT-Large with focal loss achieves 69.6\% held-out accuracy, 68.8\% weighted F1, and 83.9\% precision on the minority Multi-Year Ice class. These results show that focal-loss training offers a more useful precision--recall trade-off than weighted cross-entropy for rare ice classes and establishes a cleaner baseline for future multimodal fusion with optical, thermal, or meteorological data.
\end{abstract}


\section{Introduction}
The Arctic is warming at nearly four times the global average, a phenomenon known as Arctic amplification. The rapid decline in sea ice extent has profound implications for the global climate system and has simultaneously opened new maritime routes for shipping and resource extraction. Operational ice charting, crucial for navigation safety, currently relies on manual analysis of satellite imagery---a labor-intensive process that struggles to scale with the increasing frequency of satellite acquisitions.

\begin{figure}[htbp]
\centerline{\includegraphics[width=0.9\columnwidth]{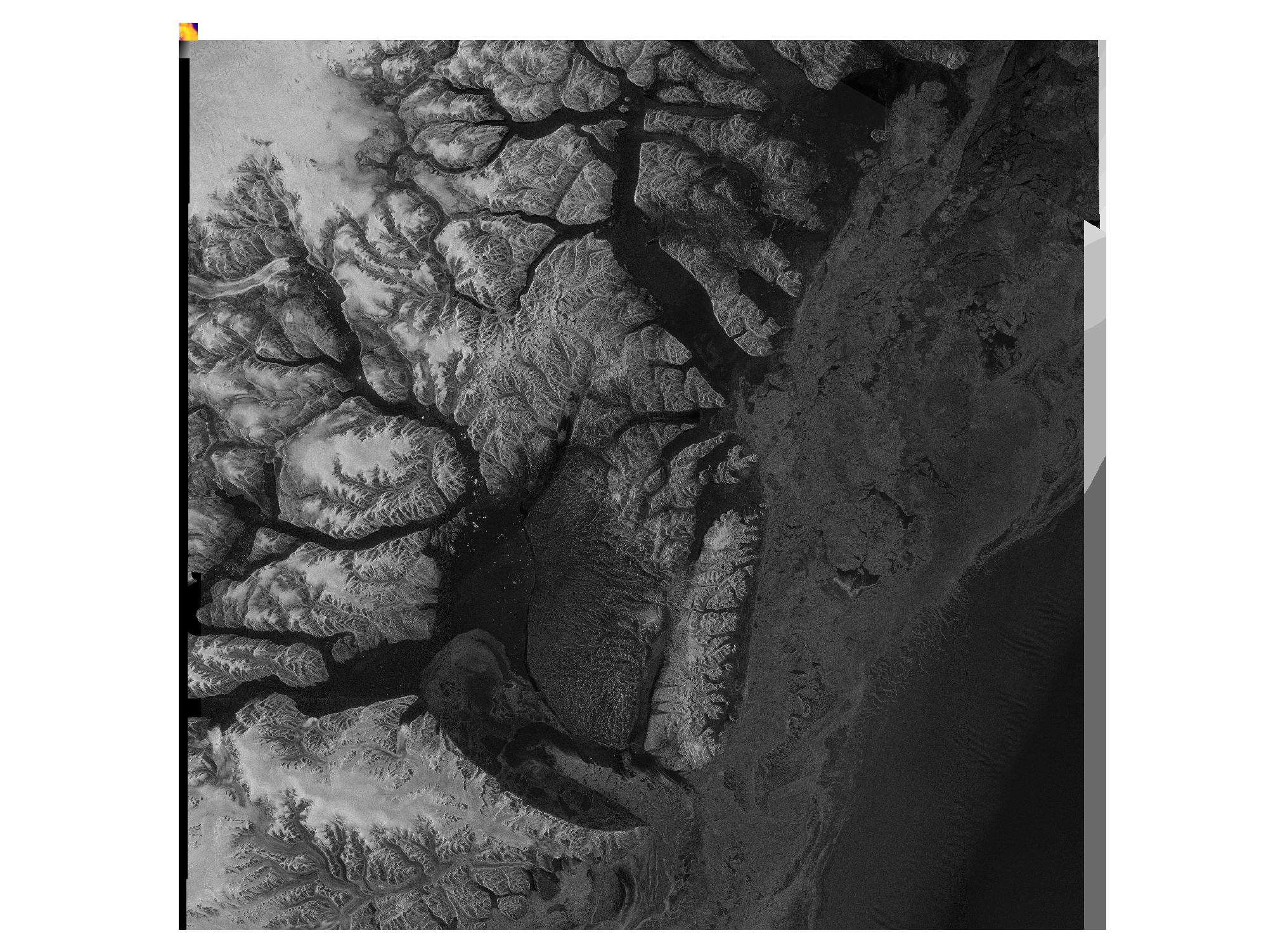}}
\caption{Sentinel-1 SAR imagery showing the ambiguity of ice signatures. Texture and context are often more discriminative than pixel intensity alone.}
\label{fig:sar_vis}
\end{figure}

Synthetic Aperture Radar (SAR), such as that from Sentinel-1, is the primary modality for ice monitoring due to its ability to image through clouds and polar darkness. However, SAR backscatter is highly ambiguous; different ice types (e.g., Young Ice vs. Multi-Year Ice) can exhibit similar intensity profiles depending on surface roughness and incidence angle \cite{zakhvatkina2019satellite} (Fig. \ref{fig:sar_vis}).


Transformers excel at capturing long-range global context \cite{dosovitskiy2020image}, which is valuable for disambiguating local SAR descriptors. Multimodal fusion with optical, thermal, or meteorological context is a promising next step for sea ice mapping, but a credible fusion study first requires a reliable SAR baseline and a leakage-safe evaluation pipeline. Accordingly, this paper focuses on establishing that baseline rather than claiming a validated multimodal system.

This paper makes two key contributions:
\begin{enumerate}
    \item A \textbf{data-centric SAR baseline} for sea ice classification using full-resolution AI4Arctic / ASIP data, leakage-aware splitting, stage-of-development labels, and dataset-specific normalization.
    \item An empirical comparison of three tested ViT configurations under class imbalance, showing that \textbf{ViT-Large + focal loss} provides the best overall accuracy and the highest precision on the Multi-Year Ice minority class.
\end{enumerate}

\section{Related Work}
Deep learning has revolutionized remote sensing analysis. While CNNs have been the standard, Vision Transformers (ViTs) are gaining traction for their ability to model global context. Aleissaee et al. \cite{aleissaee2023} highlighted the superior performance of Transformers in hyperspectral and SAR domains. Similarly, Zhang et al. \cite{zhang2024cnn} proposed a hybrid CNN-Transformer network (SI-CTFNet) to combine local feature extraction with global semantic modeling. Very recently, Xia et al. \cite{xia2025vision} successfully applied ViTs to classify broad sea surface phenomena in SAR imagery, demonstrating their superiority over CNNs for structural features. Hierarchical approaches, such as the CNN pipeline proposed by Chen et al. \cite{chen2023sea}, have also shown that decomposing the task (e.g., Ice/Water separation followed by type classification) improves performance.

Multimodal fusion is emerging as a critical frontier. Sun et al. \cite{sun2019monitoring} and Li et al. \cite{li2021fusion} explored early combinations of optical and SAR data for broad ice monitoring, often using pixel-level or mathematical fusion. More recently, Wiehle et al. \cite{wiehle2024sea} demonstrated the benefits of fusing Sentinel-1 (SAR) and Sentinel-3 (Optical/Thermal) data using CNNs. In 2025, de Lo\"e et al. \cite{deloe2025fusing} further validated this direction by showing that fusing VIIRS Ice Surface Temperature (IST) with SAR significantly aids in resolving ambiguous signatures. Addressing the critical need for accurate sea ice monitoring in the context of climate change, Ristea et al. (2023) demonstrate the efficacy of deep learning-based semantic segmentation applied to Synthetic Aperture Radar (SAR) data \cite{10283427}. The authors introduce a hybrid Convolutional Transformer Network designed to overcome the inherent limitations of traditional CNNs, which often struggle to capture long-range dependencies. By successfully integrating the global context capabilities of Transformers with the local feature extraction strengths of convolutional layers, this architecture achieves robust sea ice classification and precise atmospheric measurements, providing a significant advancement for complex remote sensing analysis.

Building upon their previous work on hybrid architectures, Ristea et al. (2024) further refined sea ice segmentation techniques by introducing a Multi-Head Transposed Attention (MHTA) Transformer. While their earlier 2023 model established the benefit of combining convolutional layers with global Transformer context, this subsequent research focuses on optimizing the attention mechanism to better handle the high-dimensional nature of SAR imagery. By utilizing transposed attention, the authors achieve a more computationally efficient segmentation process without sacrificing the model's ability to distinguish complex ice patterns, further advancing the state-of-the-art in deep learning-based remote sensing \cite{10640437}.

\subsection{Multimodal Fusion as a Future Direction}
Recent reviews in 2024 and 2025 highlight a clear shift towards multimodal data fusion to resolve the inherent ambiguities in SAR imagery \cite{andersson2025deep,li2024advancing}. For instance, recent pan-Arctic studies have shown that combining SAR with passive microwave radiometer data can improve sea ice concentration retrieval \cite{asip2024}. Comparative work on input selection has also underscored the importance of controlled ablation testing when claiming multimodal gains \cite{chen2024comparative}. Additionally, recent studies have demonstrated the efficacy of fusing co-located optical and SAR imagery \cite{chen2024colocated}. These works motivate our broader research direction, but the present paper restricts experiments to SAR-only inputs in order to establish a clean baseline first.

\section{Methodology}

\subsection{Dataset and Protocol Summary}
We use the \textbf{AI4Arctic / ASIP Sea Ice Dataset (v2)} \cite{saldo2021ai4arctic}, which contains 461 Sentinel-1 scenes from 2018--2019 matched with expert ice charts from the Greenland Ice Service. Because the goal of this paper is to establish a trustworthy SAR foundation, all experiments are restricted to the SAR modality.

\begin{table}[htbp]
\caption{Protocol Summary}
\begin{center}
\begin{tabular}{|l|l|}
\hline
\textbf{Setting} & \textbf{Value} \\
\hline
Dataset & AI4Arctic / ASIP v2 \\
Coverage & 461 scenes, 2018--2019 \\
Inputs & Sentinel-1 EW SAR HH/HV only \\
Native resolution & 40 m per pixel \\
Labels & SIGRID-3 stage of development \\
Split & Stratified patch-based, no overlap \\
Focus metrics & Accuracy, weighted F1, MYI precision/recall \\
\hline
\end{tabular}
\label{tab:protocol}
\end{center}
\end{table}

\subsection{Data-Centric Pipeline}
We implement a data-centric pipeline designed to make the SAR baseline itself reliable:

\subsubsection{Full-Resolution Input}
We utilize the original Sentinel-1 Extra Wide (EW) swath data with a pixel spacing of 40m ($10,723 \times 10,393$ pixels per scene), preserving fine-scale texture details often lost in downsampled 80m versions.

\subsubsection{Correcting Data Leakage}
Standard random splitting violates spatial autocorrelation. We implemented a \textbf{stratified, patch-based split} ensuring no spatial overlap between training and validation sets while maintaining identical class distributions.

\subsubsection{High-Fidelity Labeling}
We utilized \textbf{SIGRID-3 "Stage of Development" (SA) codes} instead of simple ice concentration, mapping them to physically distinct ice classes (e.g., New Ice, First-Year Ice, Old Ice).

\subsubsection{Dynamic Normalization}
We replaced generic ImageNet normalization with a dynamic calculation of mean and standard deviation derived directly from the training corpus.

\subsection{Model Configurations}
We evaluate three SAR-only Vision Transformer configurations under class imbalance:
\begin{itemize}
    \item \textbf{ViT-Base + Cross-Entropy (CE)}
    \item \textbf{ViT-Base + Weighted Cross-Entropy (W-CE)}
    \item \textbf{ViT-Large + Focal Loss} ($\gamma=2.0$) \cite{lin2020focal}
\end{itemize}
These experiments should be interpreted as baseline comparisons among the tested configurations, not as a full ablation of all architecture and loss combinations.

\section{Experiments and Results}
We evaluate the three SAR-only ViT configurations on the leakage-aware held-out split described above.

\subsection{Experimental Setup}
The evaluation emphasizes both overall classification quality and minority-class behavior. In addition to overall accuracy, we track weighted F1 and the precision/recall trade-off for \textit{Multi-Year Ice (MYI)}, which is the most operationally critical minority class in our experiments.

\subsection{Results}
The performance metrics of the tested Vision Transformer configurations are summarized in Table \ref{tab:results}. Among the tested configurations, \textbf{ViT-Large + Focal Loss} achieved the highest held-out accuracy and the highest MYI precision. The best model also obtained a weighted F1-score of 68.8\%.

\begin{table}[htbp]
\caption{Baseline SAR Model Performance}
\begin{center}
\begin{tabular}{|l|c|c|c|}
\hline
\textbf{Model Config} & \textbf{Acc} & \textbf{Recall (MYI)*} & \textbf{Prec. (MYI)*} \\
\hline
ViT-Base (CE) & 64.2\% & 6.6\% & 42.1\% \\
ViT-Base (W-CE) & 66.3\% & \textbf{84.8\%} & 37.2\% \\
\textbf{ViT-Large (Focal)} & \textbf{69.6\%} & 40.6\% & \textbf{83.9\%} \\
\hline
\multicolumn{4}{l}{\footnotesize *MYI: Multi-Year Ice (Critical Minority Class)}
\end{tabular}
\label{tab:results}
\end{center}
\end{table}

The experimental results highlight a significant challenge in SAR remote sensing: the severe class imbalance of Multi-Year Ice (MYI). The baseline \textit{ViT-Base (CE)} model fails to adequately capture the minority class, yielding a negligible recall of 6.6\%. While the \textit{Weighted Cross-Entropy (W-CE)} approach successfully boosts MYI detection to a recall of 84.8\%, it introduces substantial noise, resulting in a low precision of 37.2\%. Among the tested configurations, \textbf{ViT-Large + Focal Loss} achieves the most reliable balance for operational use. It reaches the highest overall accuracy of 69.6\% and a superior precision of 83.9\% for the MYI class. These results suggest that focal-loss training provides a better precision--recall trade-off than weighted cross-entropy for rare ice classes in this SAR setting. They also remain consistent with prior work showing the value of global context in sea ice segmentation \cite{10283427,10640437}.

\subsection{Discussion}
The experimental results highlight a fundamental trade-off between sensitivity and reliability in sea ice segmentation. As shown in Table \ref{tab:results}, the Weighted Cross-Entropy (W-CE) approach adopts a "Safety First" strategy, maximizing MYI recall at the expense of a 37.2\% precision rate. For operational deployment, however, the ViT-Large (Focal Loss) model is more attractive because its 83.9\% MYI precision sharply reduces false alarms in critical navigation zones.

\begin{figure}[htbp]
\centerline{\includegraphics[width=0.85\columnwidth]{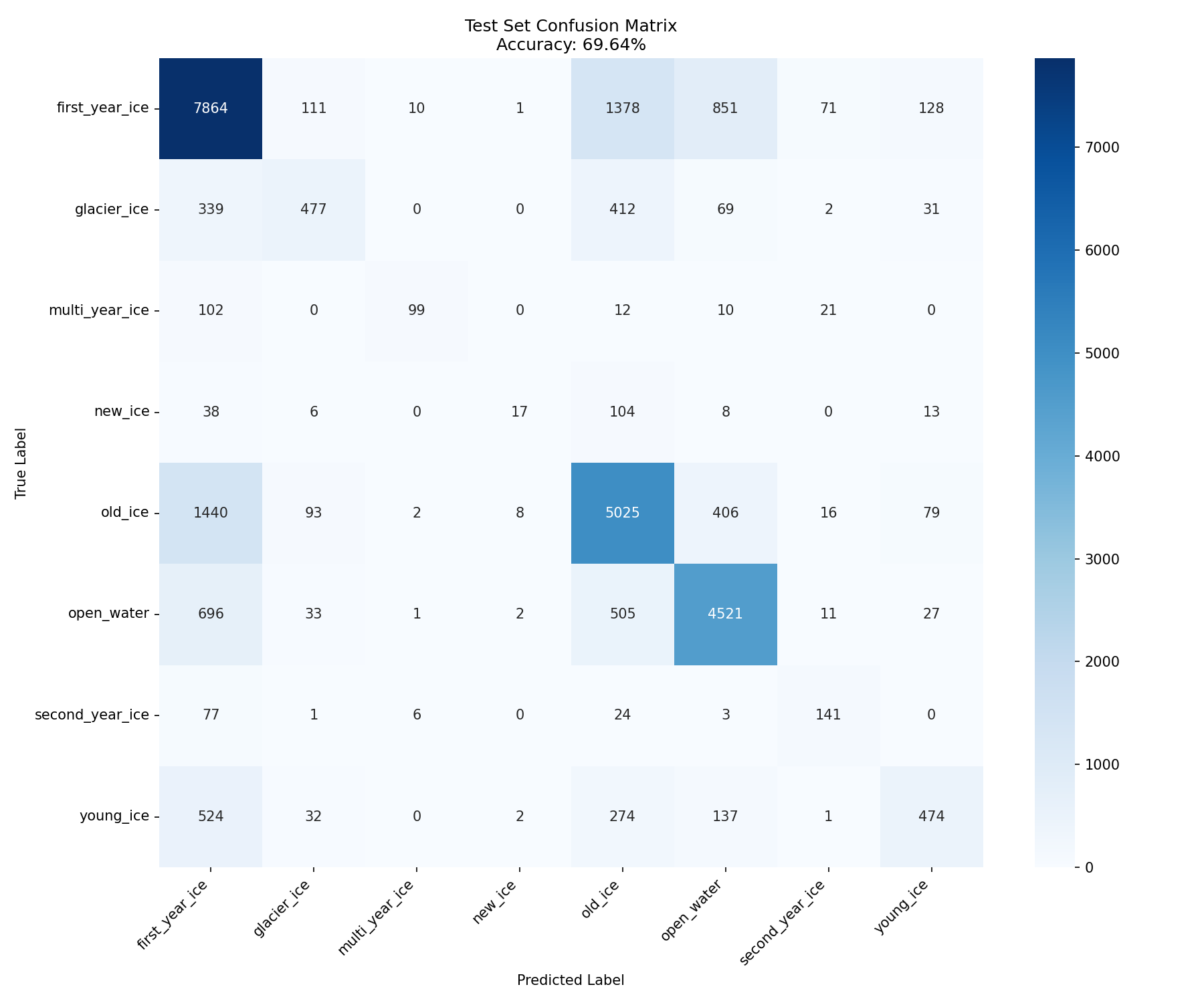}}
\caption{Confusion Matrix for the Champion ViT-Large Model.}
\label{fig:cm}
\end{figure}

An in-depth look at the Confusion Matrix (Fig. \ref{fig:cm}) reveals that the primary source of error is the morphological similarity between ice types. Significant confusion exists between First-Year Ice and Old Ice, as well as Glacier Ice being misclassified as First-Year Ice (339 instances). These errors align with the challenges noted by Ristea et al. (2023) regarding the capture of long-range dependencies in SAR data. While their work utilized Convolutional Transformer Networks to bridge this gap, and their 2024 study optimized efficiency via Multi-Head Transposed Attention, our analysis suggests that SAR-only features still struggle with several physically similar stage-of-development classes.
The confusion between Young Ice and First-Year Ice (524 instances) particularly motivates future multimodal fusion. A stronger next step is to test whether optical, thermal, passive-microwave, or meteorological context can reduce these specific ambiguities while preserving the high precision established by the SAR baseline.

\section{Conclusion}
We have presented a data-centric SAR baseline for sea ice classification using Vision Transformers and the AI4Arctic / ASIP dataset. By focusing on full-resolution SAR input, leakage-aware splitting, stage-of-development labels, and dataset-specific normalization, we establish a cleaner foundation for subsequent model development. Among the tested configurations, ViT-Large with focal loss produced the highest held-out accuracy and the highest precision on the Multi-Year Ice minority class. Future work will extend this baseline with controlled multimodal fusion experiments using optical, thermal, passive-microwave, or meteorological inputs and will evaluate whether those additions reduce the class confusions identified here.

\section{Acknowledgments}
The authors would like to thank the AI4Arctic initiative and the Automated Sea Ice Products (ASIP) team for providing the sea ice dataset used in this research. We acknowledge the European Space Agency (ESA) for making Sentinel-1 imagery freely available through the Copernicus program. This research was supported by the computational resources and facilities at The University of Texas Permian Basin.

\bibliographystyle{IEEEtran}
\balance
\bibliography{references}

\end{document}